\documentclass[10pt,onecolumn,letterpaper]{article}

\usepackage{cvpr}
\usepackage{times}
\usepackage{epsfig}
\usepackage{graphicx}
\usepackage{amsmath}
\usepackage{amssymb}
\usepackage{multirow}
\usepackage{subfig}

\newcommand{\Tr}{\mathrm{T}}

\newcommand{\RR}{{\mathbb{R}}}

\newcommand{\mm}[1]{\mathbf{#1}}

\cvprfinalcopy % *** Uncomment this line for the final submission

\begin{document}

%%%%%%%%% TITLE
%\title{Cross-version metadata mapping in large-scale video collections}
%\title{Affine-invariant diffusion geometry}
\title{Affine-invariant diffusion geometry
        for the analysis of deformable 3D shapes}

\author{Dan Raviv\\
Dept. of Computer Science\\
Technion, Israel\\
{\tt\small darav@cs.technion.ac.il}
\and
Alexander M. Bronstein\\
Dept. of Electrical Engineering\\
Tel Aviv University, Israel\\
{\tt\small bron@eng.tau.ac.il}
\and
Michael M. Bronstein\\
Dept. of Informatics\\
Universit{\`a} della Svizzera Italiana\\
Lugano, Switzerland\\
{\tt\small bronstein@ieee.org}
\and
Ron Kimmel\\
Dept. of Computer Science\\
Technion, Israel\\
{\tt\small ron@cs.technion.ac.il}
\and
Nir Sochen\\
Dept. of Applied Mathematics\\
Tel Aviv University, Israel\\
{\tt\small sochen@math.tau.ac.il}
}

\maketitle
%\thispagestyle{empty} % *** Uncomment this line for the final submission

%%%%%%%%% ABSTRACT

\begin{abstract}

We introduce an (equi-)affine invariant diffusion geometry by
 which surfaces that go through squeeze and shear transformations
 can still be properly analyzed.
The definition of an affine invariant metric enables us to construct
 an invariant Laplacian from which local and global geometric structures
are extracted.
Applications of the proposed framework demonstrate its power in
 generalizing and enriching the existing set of
 tools for shape analysis.
\end{abstract}

\section{Introduction}
\label{sec:intro}

{\em Diffusion geometry} is an umbrella term referring to geometric analysis of diffusion or random walk processes.
Such methods, first introduced in theoretical geometry \cite{berard} have matured into practical applications in the fields of manifold learning
 \cite{diff} and shape analysis \cite{levy2006lbe}.
In the shape analysis community, diffusion geometry methods were used to define low-dimensional representations for
 manifolds \cite{diff,rustamov2007lbe}, build intrinsic distance metrics and
 construct shape distribution descriptors \cite{rustamov2007lbe,MS_GMOD}, %,BroBB1},
 define spectral signatures \cite{reuter} (shape-DNA), and local descriptors
 \cite{sunHKS}. %,BroSI:HKS} % add after acceptance
  %and bags of features \cite{BroBroOvsGui09}. % add after acceptance
Diffusion embeddings were used for finding correspondence between shapes \cite{MHKCB} and detecting intrinsic symmetries \cite{ovsjanikov2008gis} .
In many settings, the construction of diffusion geometry boils down to the definition of a {\em Laplacian operator}.
Such an operator should possess certain invariance properties desired in a specific application.

In this paper, we construct {\em (equi-)affine-invariant} diffusion geometry for 3D shapes. Affine invariance is important in many applications in
the analysis of images \cite{cordelia} and 3D shapes \cite{dinosaurs}.
We first construct an affine-invariant Riemannian metric that allows us define an affine-invariant Laplacian, with which we
in turn define affine invariant diffusion geometry for surfaces.
This new geometry enables efficient computational tools that
 handle both non-rigid approximately-isometric deformations of the surface
 together with equi-affine transformations of the embedding space.
We demonstrate the usefulness of our construction in a range of shape analysis
 applications, such as retrieval, correspondence, and symmetry detection.

\section{Background}
\label{sec:backgr}

% Please disable WRAP!
Let $X$ denote a compact two-dimensional Riemannian manifold (possibly with boundary) representing the outer
 boundary of a physical solid object in the 3D space.
The {\em Riemannian metric tensor} $g$ is defined as a local inner product $\langle \cdot, \cdot \rangle_x$
 on the {\em tangent plane} $T_x X$ at each point $x\in X$.
Given a smooth scalar field $f : X \rightarrow \mathbb{R}$, its {\em gradient} $\nabla f$ at point $x$ is defined through
 the relation $f(x+dr) = f(x) + \langle \nabla f(x), dr \rangle_x$,
 where $dr \in T^*_x X$ is an infinitesimal tangent vector.
The positive semi-definite self-adjoint {\em Laplace-Beltrami operator} $\Delta_g$ associated with the metric
  tensor $g$ is defined by the identity
\begin{eqnarray}
\int f \Delta_g h \, da &=& - \int \langle \nabla f, \nabla h\rangle_x da
\label{eq:lbo}
\end{eqnarray}
holding for any pair of smooth scalar fields $f,h:X\rightarrow \mathbb{R}$; here $da$ denotes the standard area measure on $X$.
Whenever possible, we will omit the subscript $g$ and refer to the Laplace-Beltrami operator simply as to $\Delta$.

Assuming further that the manifold is embedded isometrically in $\mathbb{R}^3$ and (possibly, locally) parametrized by a
 regular map $\mm{x} : U \rightarrow \RR^3$ over a planar domain $U$, the metric tensor $g$ assumes
 the form of a $2\times 2$ positive-definite matrix called the \emph{first fundamental form}, whose elements are given by the inner products
%\begin{eqnarray}
%\mm{G} &=& \left(\frac{\partial \mm{x}}{\partial \mm{u}}\right)^\Tr \frac{\partial \mm{x}}{\partial \mm{u}}
$g_{ij} = \frac{\partial \mm{x}^\Tr}{\partial u_i}  \frac{\partial \mm{x} }{\partial u_j}$.
%\end{eqnarray}
%  $g_{ij} = \langle \frac{\partial}{\partial u_i} X, \frac{\partial}{\partial u_j} X \rangle$,
%Here, $\frac{\partial \mm{x}}{\partial \mm{u}}$ denotes the $3 \times 2$ Jacobian matrix of $\mm{x}$ w.r.t. to the
%vector of coordinates $\mm{u}$, whose $ij$-th element is given by $\frac{\partial x_i}{\partial u_j}$.

The Laplace-Beltrami operator can be expressed in the parametrization coordinates as
%\begin{eqnarray}
%\Delta f = - \frac{1}{\sqrt{\det \mm{G}}} \frac{\partial}{\partial \mm{u}^\Tr} \left(\det \mm{G} \cdot \mm{G}^{-1} \frac{\partial f}{\partial \mm{u}}\right) %\partial_i \sqrt{g} g^{ij}\partial_j,
%\label{eq:lbo1}
%\end{eqnarray}
%
\begin{eqnarray}
 \Delta & = & - \frac{1}{\sqrt{g}} \partial_i \sqrt{g} g^{ij}\partial_j,
\label{eq:lbo1}
\end{eqnarray}
where we use Einstein's summation convention, according to which $g = \mathrm{det}(g_{ij})$ denotes
the determinant of the metric, $g^{ij}$ are the components of the inverse metric tensor,
and repeated indices are summed over.
When the metric is Euclidean ($g_{ij} = \mm{I}$), the  operator reduces to the
 familiar
% \begin{eqnarray}
$\Delta f = - \left(\frac{\partial^2 f}{\partial u^2_1} + \frac{\partial^2 f}{\partial u^2_2}\right)$
%\end{eqnarray}
(note that we define the Laplacian with the minus sign
 in order to ensure its positive semi-definiteness).

%Here, we define a new metric tensor $g$ that is invariant to volume preserving affine
% transformations of the embedding space $\mathrm{R}^3$.
%It would allow us to define new geometries for shape analysis.
%In this note we focus on the invariant {\em diffusion geometry} that is constructed from an
% invariant Laplace Beltrami opertor.

%Diffusion geometry arises from the {\em heat equation},
The Laplace-Beltrami operator gives rise to the partial differential equation
\begin{eqnarray}
\left(\frac{\partial}{\partial t} + \Delta \right)f(t,x) = 0,
\label{eq:heat}
\end{eqnarray}
called the \emph{heat equation}. The heat equation describes
the propagation of heat on the surface and its solution $f(t,x)$ is the heat distribution at a point $x$ in time $t$.
The initial condition of the equation is some initial heat distribution $f(0,x)$;
if $X$ has a boundary, appropriate boundary conditions must be added.
The solution of~(\ref{eq:heat}) corresponding to a point initial condition
 $f(0,x) = \delta(x-x')$, is called the {\em heat kernel} and represents the amount of heat
 transferred from $x$ to $x'$ in time $t$ due to the diffusion process.
Using spectral decomposition, the heat kernel can be represented as
\begin{eqnarray}
h_t(x,x') &=& \sum_{i\geq 0} e^{-\lambda_i t} \phi_i(x) \phi_i(x')
\end{eqnarray}
where $\phi_i$ and $\lambda_i$ are, respectively, the eigenfunctions and eigenvalues of the Laplace-Beltrami operator
 satisfying $\Delta \phi_i = \lambda_i \phi_i$ (without loss of generality, we assume $\lambda_i$ to be sorted in increasing order starting with
 $\lambda_0 = 0$).
Since the Laplace-Beltrami operator is an {\em intrinsic} geometric quantity, i.e., it can be expressed solely in terms of the
 metric of $X$, its eigenfunctions and eigenvalues as well as the heat kernel are invariant under isometric transformations of
 the manifold.

The value of the heat kernel $h_t(x,x')$ can be interpreted as the transition probability
density of a random walk of length $t$ from the point $x$ to the point $x'$.
This allows to construct a family of intrinsic metrics known as \emph{diffusion metrics},
\begin{eqnarray}
d^2_t(x,x') = \int \left( h_t(x,\cdot) - h_t(x',\cdot) \right)^2 da = \sum_{i> 0} e^{-\lambda_i t} (\phi_i(x) - \phi_i(x'))^2,
\label{eq:diffdist}
\end{eqnarray}
which measure the ``connectivity rate'' of the two points by paths of length $t$.

The parameter $t$ can be given the meaning of \emph{scale}, and the family $\{ d_t \}$ can be thought of as a scale-space of metrics.
By integrating over all scales, a \emph{scale-invariant} version of~(\ref{eq:diffdist}) is obtained,
\begin{eqnarray}
d_\mathrm{CT}^2(x,x')  &=&
    2\int_0^\infty d_t^2(x,x') dt = \sum_{i> 0} \frac{1}{\lambda_i} (\phi_i(x) - \phi_i(x'))^2.
    \label{eq:ct-dist}
\end{eqnarray}
This metric is referred to as the {\em commute-time distance} and can be interpreted as the connectivity rate by paths of any length.
We will broadly call constructions related to the heat kernel, diffusion and commute time metrics as \emph{diffusion geometry}.

\section{Affine-invariant diffusion geometry}
\label{sec:affine}

An \emph{affine transformation} $\mm{x} \mapsto \mm{A}\mm{x} + \mm{b}$ of the three-dimensional Euclidean space
can be parametrized by a regular $3 \times 3$ matrix $\mm{A}$ and a $3\times 1$ vector $\mm{b}$ since all constructions discussed here
are trivially translation invariant, we will omit the vector $\mm{b}$.
The transformation is called \emph{special affine} or \emph{equi-affine} if it is volume-preserving, i.e., $\det \mm{A} = 1$.

As the standard Euclidean metric is not affine-invariant, the Laplace-Beltrami operators associated with $X$ and $\mm{A} X$ are generally distinct, and so are
the resulting diffusion geometries.
In what follows, we are going to substitute the Euclidean metric by its equi-affine invariant counterpart. That, in turn, will induce an equi-affine-invariant Laplace-Beltrami operator and define equi-affine-invariant diffusion geometry.
%With some abuse of terminology, we will, whenever possible, refer to equi-affine-invariant quantities simply
%as to ``affine-invariant''.

The equi-affine metric can be defined through the parametrization of a curve \cite{Axioms:Image:Proc:morel93,Proj:Inv:Bruck97,Buchin1983AffineGeometry,Best:Scale:Space:Caselles:96,Geometric:Evolutions:Olver:97,Sochen04}.
Let $C$ be a curve on $X$ parametrized by $p$.
By the chain rule,
\begin{eqnarray}
\frac{dC}{dp} &=& \mm{x}_1 \frac{d u_1}{d p} + \mm{x}_2 \frac{d u_2}{d p} \nonumber\\
\frac{d^2C}{dp^2} &=& \mm{x}_1 \frac{d^2 u_1}{dp^2} + \mm{x}_2 \frac{d^2 u_2}{dp^2} + \mm{x}_{11} \left(\frac{du_1}{dp}\right)^2 +  2\mm{x}_{12}\frac{du_1}{dp} \frac{du_2}{dp}+\mm{x}_{22} \left(\frac{du_2}{dp}\right)^2,
\label{eq:chain-rule}
\end{eqnarray}
where, for brevity, we denote $\mm{x}_i = \frac{\partial \mm{x}}{\partial u_i}$ and $\mm{x}_{ij} = \frac{\partial^2 \mm{x}}{\partial u_i \partial u_j}$.
As volumes are preserved under the equi-affine group of transformations,
 we define the invariant arclength $p$ through
\begin{eqnarray}
\label{eq:arclength}
 \det (\mm{x}_1,\mm{x}_2,C_{pp}) &=& 1.
\end{eqnarray}
Plugging (\ref{eq:chain-rule}) into (\ref{eq:arclength}) yields
\begin{eqnarray}
\label{eq:arclength_1}
dp^2 =  \det (\mm{x}_1,\mm{x}_2, \mm{x}_{11} du_1^2 + 2\mm{x}_{12}du_1 du_2+\mm{x}_{22}du_2^2),
\end{eqnarray}
from where we readily have an equi-affine-invariant pre-metric tensor
\begin{eqnarray}
 \label{eq:equi_metric}
  \hat{g}_{ij} &=& \tilde{g}_{ij}  \left|\tilde{g}\right|^{-1/4},
\end{eqnarray}
where  $\tilde{g}_{ij} = \det (\mm{x}_1,\mm{x}_2,\mm{x}_{ij})$.
The pre-metric tensor~(\ref{eq:equi_metric})  defines a true metric only on strictly
 convex surfaces \cite{Buchin1983AffineGeometry}; in more general cases, it might cease
 from being positive definite.
In order to deal with arbitrary surfaces, we extend the metric definition by restricting the eigenvalues of the tensor to be positive.
Representing $\hat{g}$ as a $2 \times 2$ matrix admitting the eigendecomposition $\hat{\mm{G}} = \mm{U} \mm{\Gamma}\mm{U}^\Tr$,  where
$\mm{U}$ is orthonormal and $\mm{\Gamma} = \mathrm{diag}\{\gamma_1,\gamma_2 \}$, we compose a new first fundamental form matrix
$\mm{G} = \mm{U} |\mm{\Gamma}| \mm{U}^\Tr$. The corresponding metric tensor $g$ is positive definite and is equi-affine invariant.

\begin{figure*}[tp]

  \centering
\includegraphics[width=0.85\linewidth]{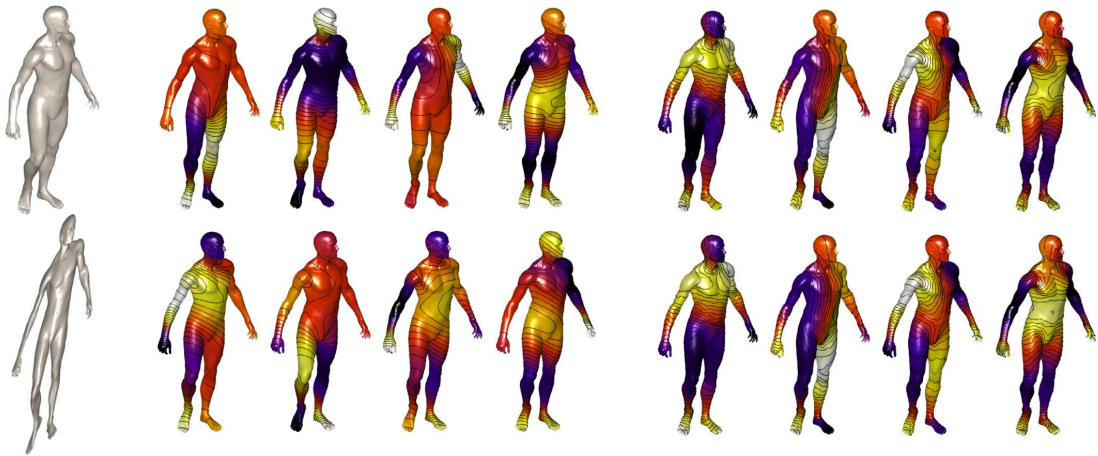}
  \caption{\label{Fig:eigenfunctions} \small Four eigenfunctions of the standard (second through fifth columns) and the proposed
  equi-affine-invariant (four rightmost columns)  Laplace-Beltrami operators. Two rows show a shape and its equi-affine transformation.
  For convenience of visualization, eigenfunctions are overlaid onto the untransformed shape.}

\end{figure*}

\begin{figure}[tp]

  \centering
\includegraphics[width=1\columnwidth]{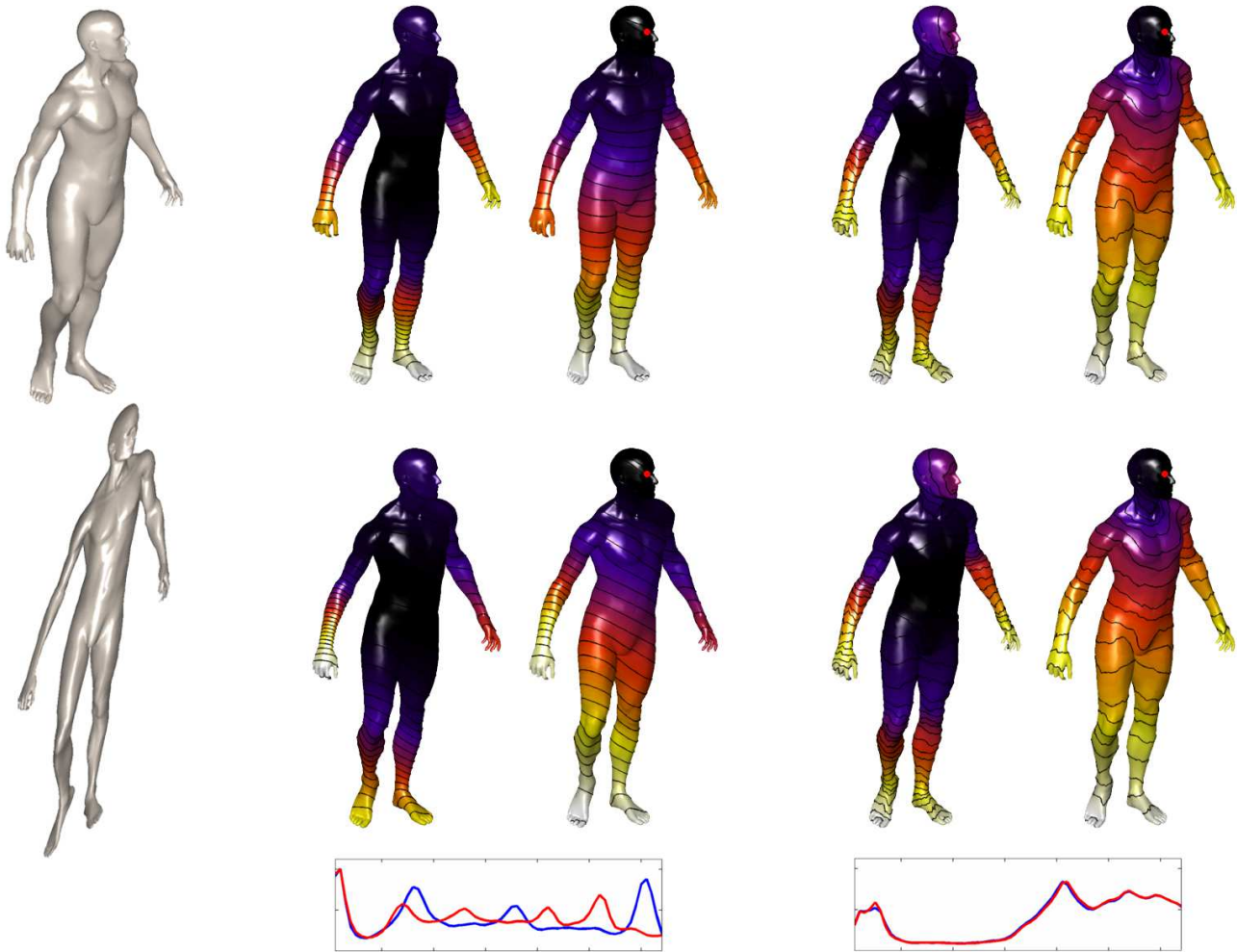}
  \caption{\label{Fig:dist-kern} \small Heat kernel signature $h_t(x,x)$ and diffusion metric ball (second and third columns, respectively), and their
  affine-invariant counterparts (fourth and fifth columns, respectively). Two rows show a shape and its equi-affine transformation.
  For convenience of visualization, the kernel and the metric are overlaid onto the untransformed shape. Plots under the figure show the corresponding
  metric distributions before and after the transformation. }

\end{figure}

Plugging this $g$
 into (\ref{eq:lbo}), we obtain an equi-affine-invariant Laplace-Beltrami operator $\Delta_g$.
Such an operator defines an equi-affine-invariant diffusion geometry, i.e., the eigenfunction, heat kernel, and diffusion metrics
it generates remain unaltered by a global volume-preserving affine transformation of the shape (Figures~\ref{Fig:eigenfunctions}--\ref{Fig:dist-kern}).
%
%Because of its scale invariance property, the commute time metric induced by the
%equi-affine-invariant $\Delta_{\hat{g}}$ is affine-invariant.

\section{Discretization}
\label{sec:num}

\begin{figure}[tp]
  \centering \includegraphics[width=\columnwidth]{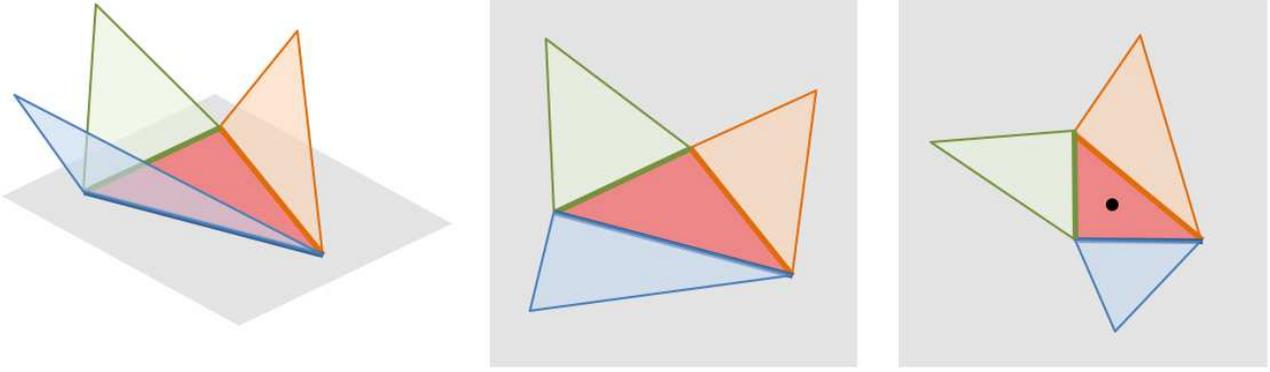}
   \caption{\label{fig:unfolding}
\small{   Left to right: part of a triangulated surface about a specific triangle.
   The three neighboring triangles together with the central one are unfolded flat to the plane.
   The central triangle is canonized into a right isosceles triangle; three
   neighboring triangles follow the same planar affine transformation.
   Finally, the six surface coordinate values at the vertices are used to interpolate
   a quadratic surface patch from which the metric tensor is computed. }
   }
\end{figure}

In order to compute the equi-affine metric we need to evaluate the second-order
 derivatives of the surface with respect to some parametrization coordinates.
 While this can be done practically in any representation, here we assume that the surface is
 given as a triangular mesh.
For each triangular face, the metric tensor elements are calculated from a quadratic surface
 patch fitted to the triangle itself and its three adjacent neighbor triangles.
 The four triangles are unfolded to the plane, to which an affine transformation is applied
 in such a way that the central triangle becomes a unit simplex. The coordinates of this planar
 representation are used as the parametrization $\mm{u}$ with respect to which the first fundamental form coefficients are computed
at the barycenter of the simplex (Figure \ref{fig:unfolding}).
This step is performed for every triangle of the mesh.

Having the discretized first fundamental form coefficients, our next goal is to discretize the Laplace-Beltrami operator.
Since our final goal is not the operator itself but its eigendecomposition, we skip the construction of the Laplacian
and discretize its eigenvalues and eigenfunctions directly. This is achieved using the finite elements method (FEM) proposed in \cite{Dziuk88}
 and used in shape analysis in \cite{reuter2009discrete}.
For that purpose, we translate the eigendecomposition of the Laplace-Beltrami operator
 $\Delta \phi = \lambda \phi$
 into a {\em weak form}
\begin{eqnarray}
\int \psi_k \Delta \phi \,  da &=& \lambda \int \psi_k \phi \, da
\label{eq:weakform}
\end{eqnarray}
with respect to some basis $\{\psi_k\}$ spanning a (sufficiently smooth) subspace of $L^2(X)$.
Specifically, we choose the $\psi_k$'s to be the first-order finite element function obtaining the value of one
 at a vertex $k$ and decaying linearly to zero in its $1$-ring (the size of the basis equals to the number of vertices in the mesh).
Substituting these functions into (\ref{eq:weakform}), we obtain
 \begin{eqnarray}
  \int \psi_k \Delta \phi \, da  \ = \ \int  \langle \nabla \psi_k, \nabla \phi \rangle_x \, da  = \ \int g^{ij} (\partial_i \phi) (\partial_j \psi_k)  \, da \ = \  \lambda \int \psi_k  \phi \, da.
 \end{eqnarray}
Next, we approximate the eigenfunction $\phi$ in the finite element basis by $\phi = \sum_{l=1} \alpha_l \psi_l$.
This yields
 \begin{eqnarray*}
\int g^{ij} (\partial_i \sum_l \alpha_l  \psi_l) (\partial_j \psi_k)  \, da
&  =& \lambda \int \psi_k  \sum_l \alpha_l \psi_l \, da,
\end{eqnarray*}
 or, equivalently,
\begin{eqnarray}
  \sum_l \alpha_l \int g^{ij} (\partial_i  \psi_l) (\partial_j \psi_k)  \, da
  &=& \lambda \sum_l \alpha_l \int \psi_k   \psi_l \, da. \nonumber
\end{eqnarray}
The last equation can be rewritten in matrix form as a generalized eigendecomposition
 problem $\mm{A} \mm{\alpha} = \lambda  \mm{B} \mm{\alpha}$ solved for the coefficients $\alpha_l$, where
\begin{eqnarray*}
  a_{kl} &=& \int g^{ij} (\partial_i  \psi_l) (\partial_j \psi_k)  \, da, \cr
  b_{kl} &=&  \int \psi_k   \psi_l \, da,
\end{eqnarray*}
and the local surface area is expressed in parametrization coordinates as $da = \sqrt{g}du_1 du_2$.

\section{Applications and results}
\label{sec:app}

The proposed equi-affine-invariant Laplacian is a practical tool that can be employed in the construction of local and global diffusion geometric structures
used in standard approaches in shape analysis, substituting the traditional non-invariant Laplace-Beltrami operator.
%
%In this section, we discuss the applications to shape retrieval, correspondence and shape matching, and symmetry detection.
%
In what follows, we detail the construction of such structures and show applications in shape retrieval, correspondence, matching, and symmetry detection.

\subsection{Shape retrieval}

Sun \emph{et al.} \cite{sunHKS} proposed using the diagonal of the heat kernel, $h_t(x,x)$, as a local descriptor of the shape, referred to as the {\em heat kernel signature} (HKS).
In practice, the descriptor is computed by sampling the time parameter $t$ at a discrete set of points, $t_1,\hdots,t_n$, and collecting the
corresponding values of $h_t(x,x)$ into a vector $\mathbf{p}(x) = (h_{t_1}(x,x),\hdots,h_{t_n}(x,x))$.
The HKS descriptor is isometry-invariant, easy to compute, and is provably informative \cite{sunHKS}.
%
%%A scale-invariant version (SI-HKS) can be obtained by retaining the Fourier transform magnitude of the log-transformed HKS sampled on a logarithmic time scale. %\cite{BroSI:HKS}.
%%%
%%By replacing the standard Laplacian in HKS with our equi-affine-invariant operator, an \emph{equi-affine-invariant HKS}  (EAI-HKS) is obtained,
%%while using the SI-HKS instead of HKS yields a truly \emph{affine-invariant} version of the descriptor.
The affine-invariant HKS descriptors can be used in one-to-one feature-based shape matching methods \cite{sunHKS},
or in large-scale shape retrieval applications using the {\em bags of features} paradigm \cite{BroBroOvsGui09}.
In the latter, the shape is considered as a collection of ``geometric words'' from a fixed ``vocabulary'' and is described by the statistical distribution of such words.
The vocabulary is constructed off-line by clustering the descriptor space.
Then, for each point on the shape, the descriptor is replaced by the nearest vocabulary word by means of vector quantization.
Counting the frequency of each word, a bag of features is constructed.
The similarity of two shapes is then computed as the distance between the corresponding bags of features.

To evaluate the performance of the proposed approach for the construction of local descriptors, we used the Shape Google
 framework \cite{BroBroOvsGui09}
 based on standard and equi-affine-invariant HKS.
 Both descriptors were computed at six scales ($t = 1024, 1351.2, 1782.9, 2352.5$, and $4096$). Bags of features were computed using soft vector quantization with variance taken as twice the median of
all distances between cluster centers in a vocabulary of $64$ entries. Approximate nearest neighbor method \cite{AryaMount:98:ICP} was used for vector quantization.
Both the standard and the affine-invariant Laplace-Beltrami operator discretization were computed using finite elements.
Heat kernels were approximated using the first smallest $100$ eigenpairs.
%
%The geometric vocabulary size was set to 64. % $ \approx 300K $ ``geometric words''.

\begin{figure*}[htb]
\centering \includegraphics*[width=1\linewidth]%,bb=1 1 3520 770]
                {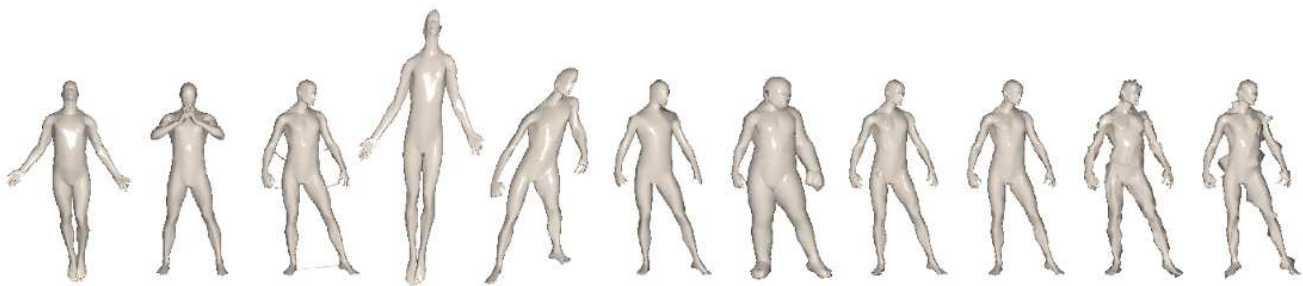}
   \caption{\label{fig:shrec} \small Examples of query shape transformations used in the shape retrieval experiment (left to right): null, isometry, topology, affine, affine+isometry, sampling, local scale, holes, microholes, Gaussian noise, shot noise.}
\end{figure*}

Evaluation was performed using the SHREC 2010 robust large-scale shape retrieval benchmark methodology. % \cite{SHRECr}.
The dataset consisted of two parts: 793 shapes from 13 shape classes with simulated transformation of different types (Figure~\ref{fig:shrec}) and strengths (60 per shape) used as queries, and the remaining 521 shapes used as the queried corpus.
Transformations classes {\em affine} and {\em isometry+affine} were added to the original SHREC query set, representing, respectively, equi-affine transformations of different strengths of the null shape and its approximate isometry.
The combined dataset consisted of 1314 shapes.
Retrieval was performed by matching 780 transformed queries to the 534 null shapes. Each query had one correct corresponding null shape in the dataset.
Performance was evaluated using precision/recall characteristic.
{\em Precision} $P(r)$ is defined as the percentage of relevant shapes in the first $r$ top-ranked retrieved shapes.
{\em Mean average precision} (mAP), defined as $mAP = \sum_r P(r) \cdot rel(r)$,
where $rel(r)$ is the relevance of a given rank, was used as a single measure of performance. Intuitively, mAP is interpreted as the area below the precision-recall curve.
Ideal retrieval performance (mAP=100\%) is achieved when all queries return relevant first matches.
Performance results were broken down according to transformation class and strength.

Tables~\ref{tab:ai}--\ref{tab:fem} show that % in contrast to
the equi-affine version of the ShapeGoogle approach
obtains slightly higher precision than the original ShapeGoogle in all SHREC'10 transformations.
We attribute this phenomenon to the smoothing effect  of the second order interpolation.
In addition, equi-affine ShapeGoogle exhibits nearly perfect retrieval under equi-affine transformations
in affine transformations, where the original approach fails.
%In addition we can see an improvement in almost all of the transformations,

\begin{table}
\centering
\begin{tabular}{lccccc}
& \multicolumn{5}{c}{\bf\small Strength} \\
\cline{2-6}
{\bf\small Transform.} & {\bf\small 1} & {\bf\small $\leq$2} & {\bf\small $\leq$3} & {\bf\small $\leq$4} & {\bf\small $\leq$5}\\
\hline
{\small\em Isometry} & {\small 100.00} & {\small 100.00} & {\small 100.00} & {\small 100.00} & {\small 99.23} \\
{\small\em Affine} & {\small 100.00} & {\small 100.00} & {\small 100.00} & {\small 100.00} & {\small 97.44} \\
{\small\em Iso.+Affine} & {\small 100.00} & {\small 100.00} & {\small 100.00} & {\small 100.00} & {\small 100.00} \\
{\small\em Topology} & {\small 96.15} & {\small 94.23} & {\small 91.88} & {\small 89.74} & {\small 86.79} \\
{\small\em Holes} & {\small 100.00} & {\small 100.00} & {\small 100.00} & {\small 100.00} & {\small 100.00} \\
{\small\em Micro holes} & {\small 100.00} & {\small 100.00} & {\small 100.00} & {\small 100.00} & {\small 100.00} \\
%{\small\em Scale} & {\small 0.60} & {\small 29.82} & {\small 26.88} & {\small 20.99} & {\small 17.37} \\
{\small\em Local scale} & {\small 100.00} & {\small 100.00} & {\small 94.74} & {\small 82.39} & {\small 73.97} \\
{\small\em Sampling} & {\small 100.00} & {\small 100.00} & {\small 100.00} & {\small 96.79} & {\small 86.10} \\
{\small\em Noise} & {\small 100.00} & {\small 100.00} & {\small 89.83} & {\small 78.53} & {\small 69.22} \\
{\small\em Shot noise} & {\small 100.00} & {\small 100.00} & {\small 100.00} & {\small 97.76} & {\small 89.63} \\
%{\small\em Mixed} & {\small 34.12} & {\small 40.08} & {\small 30.83} & {\small 24.16} & {\small 19.63} \\
%\hline
%{\small{\bf Average}} & {\small 84.09} & {\small 84.66} & {\small 83.39} & {\small 81.18} & {\small 78.33} \\
\hline
\end{tabular}
\caption{\small Performance (mAP in \%) of Shape Google with equi-affine-invariant HKS descriptors. \label{tab:ai}}
\end{table} 
\begin{table}
\centering
\begin{tabular}{lccccc}
& \multicolumn{5}{c}{\bf\small Strength} \\
\cline{2-6}
{\bf\small Transform.} & {\bf\small 1} & {\bf\small $\leq$2} & {\bf\small $\leq$3} & {\bf\small $\leq$4} & {\bf\small $\leq$5}\\
\hline
{\small\em Isometry} & {\small 100.00} & {\small 100.00} & {\small 100.00} & {\small 100.00} & {\small 100.00} \\
{\small\em Affine} & {\small 100.00} & {\small 86.89} & {\small 73.50} & {\small 57.66} & {\small 46.64} \\
{\small\em Iso.+Affine} & {\small 94.23} & {\small 86.35} & {\small 76.84} & {\small 70.76} & {\small 65.36} \\
{\small\em Topology} & {\small 100.00} & {\small 100.00} & {\small 98.72} & {\small 98.08} & {\small 97.69} \\
{\small\em Holes} & {\small 100.00} & {\small 96.15} & {\small 92.82} & {\small 88.51} & {\small 82.74} \\
{\small\em Micro holes} & {\small 100.00} & {\small 100.00} & {\small 100.00} & {\small 100.00} & {\small 100.00} \\
%{\small\em Scale} & {\small 1.33} & {\small 41.82} & {\small 44.02} & {\small 33.75} & {\small 27.37} \\
{\small\em Local scale} & {\small 100.00} & {\small 100.00} & {\small 97.44} & {\small 87.88} & {\small 78.78} \\
{\small\em Sampling} & {\small 100.00} & {\small 100.00} & {\small 100.00} & {\small 96.25} & {\small 91.43} \\
{\small\em Noise} & {\small 100.00} & {\small 100.00} & {\small 100.00} & {\small 99.04} & {\small 99.23} \\
{\small\em Shot noise} & {\small 100.00} & {\small 100.00} & {\small 100.00} & {\small 98.46} & {\small 98.77} \\
%{\small\em Mixed} & {\small 62.99} & {\small 66.35} & {\small 56.78} & {\small 48.01} & {\small 39.12} \\
%\hline
%{\small{\bf Average}} & {\small 84.94} & {\small 85.32} & {\small 83.81} & {\small 80.80} & {\small 77.86} \\
\hline
\end{tabular}
\caption{\small Performance (mAP in \%) of Shape Google with HKS descriptors. \label{tab:fem}}
\end{table}

\subsection{Global structures}

The equi-affine-invariant Laplacian can also be employed in the construction of global geometric structures.
By plugging it into (\ref{eq:diffdist}),
a family of equi-affine-invariant diffusion distances is obtained. Similarly,
a truly affine-invariant version of the commute time metric is obtained by using the equi-affine-invariant operator in (\ref{eq:ct-dist}).
%
%
%Recent works \cite{rustamov2007lbe,MS_GMOD,BroBB1} showed that global shape descriptors can be constructed considering {\em distributions} of intrinsic distances.
%%
%Given some intrinsic distance metric $d_X$ (e.g. diffusion or commute-time), the cumulative distribution is computed as
%\begin{eqnarray}
%F_X(\alpha) &=& \int_{X\times X} \chi_{d_X(x,x') \leq \alpha} d\mu_X(x)\times \mu_X(x'),
%\end{eqnarray}
%where $\chi$ denotes an indicator function. $F_X(\alpha)$ defined this way is the measure of % pairs of points the distance between which in no larger than
%$\alpha$; $F_X(\infty)$ is the squared area of $X$.
%%
%Given two shapes $X$ and $Y$ with the corresponding distance metrics $d_X, d_Y$, the
%similarity is computed as a statistical distance (e.g. Kullback-Leibler or Bhattcharyya)
%between the distributions $F_X$ and $F_Y$ \cite{BroBB1}.
%%
%A particular choice of $d_X, d_Y$ as the diffusion distance arising from the
%affine-invariant Laplace-Beltrami operator makes such a similarity criterion
% affine-invariant; the use of commute-time distance also adds scale-invariance.
%
%
%Affine-invariant metrics
These metric structures can be used in the {\em Gromov-Hausdorff framework} \cite{gro:GEOMETRY,mem:sap1:GEOMETRY},
  in which shapes are modeled as metric spaces, and the similarity of two shapes $(X,d_X)$ and $(Y,d_Y)$  is
  established by looking at the
minimum-distortion correspondence between them. A {\em correspondence} is defined as a subset
 $\mathcal{C} \subset X\times Y$ such that for all $x\in X$ there exists $y\in Y$ such that $(x,y) \in \mathcal{C}$,
 and vice versa, for all $y\in Y$ there exists $x\in X$ such that $(x,y) \in \mathcal{C}$.
The {\em distortion} of $\mathcal{C}$ is defined as
\begin{eqnarray}
\mathrm{dis}(\mathcal{C}) &=& \max_{(x,y), (y,y')\in \mathcal{C}} |d_X(x,x') - d_Y(y,y')|.
\end{eqnarray}
The {\em Gromov-Hausdorff distance} is given as the minimum of the distortion over all possible correspondences,
\begin{eqnarray}
 d_\mathrm{GH}(X,Y) &=& \frac{1}{2} \min_{\mathcal{C}} \mathrm{dis}(\mathcal{C}),
 \label{eq:GH}
\end{eqnarray}
and serves as a criterion for shape similarity in the sense that two shapes with $d_\mathrm{GH}(X,Y) \le \epsilon$ are at most $2\epsilon$-isometric,
and, vice versa, two $\epsilon$-isometric shapes have at most $d_\mathrm{GH}(X,Y) \le 2\epsilon$.
A byproduct of this problem is the minimum-distortion correspondence $\mathcal{C}$.

%This framework can be extended by considering local structures (descriptors)
% in additional to global structures (distance metrics) \cite{thorstensen,chaohui:INRIA}.

\begin{figure}[tpb]
  \centering \includegraphics*[width=1\columnwidth]{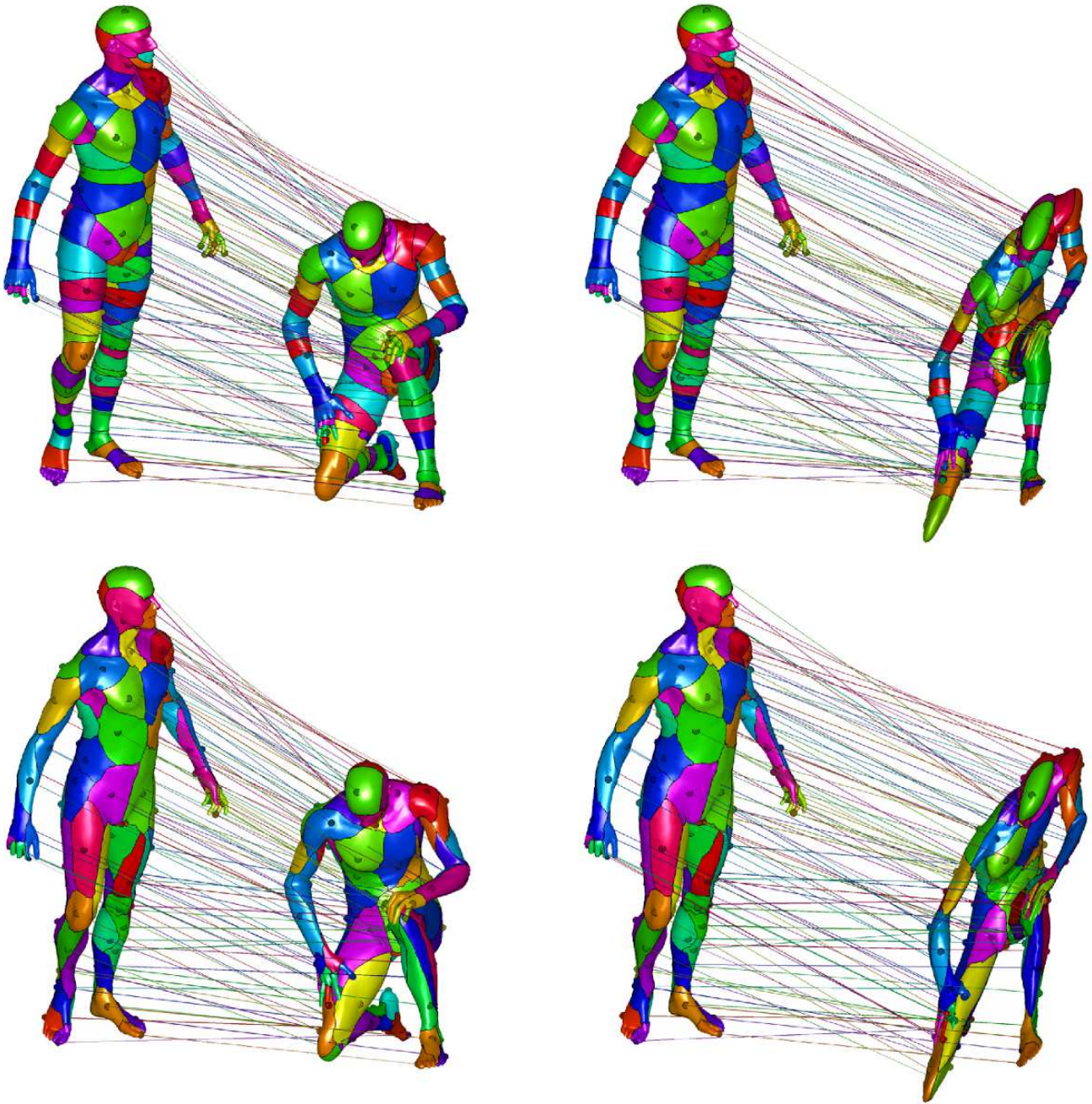}
  \caption{ \label{Fig:gmds-matching} \small The GMDS framework is used to calculate correspondence between a shape and its isometry (left) and
  isometry followed by an equi-affine transformation (right).
  Matches between shapes are depicted as identically colored Voronoi cells.
Standard diffusion distance (first row) and its equi-affine-invariant counterpart (second row) are used as the metric structure
in the GMDS algorithm. Inaccuracies obtained in the first case are especially visible in the abdominal region.
}
\end{figure}

Figure~\ref{Fig:gmds-matching} shows the correspondences obtained between an equi-affine transformation of a shape using the
standard and the equi-affine-invariant versions of the diffusion metric. Minimization of a least-squares
version of (\ref{eq:GH}) was performed using the generalized multidimensional scaling (GMDS) algorithm. %\cite{bro:bro:kim:PNAS}.
In the case of the standard diffusion metric, the embedding distortion grew by over $6$ times as the result of the transformation,
while in the case of the proposed invariant diffusion metric, the increase was by mere $16\%$.

\subsection{Intrinsic symmetry detection}

Ovsjanikov \emph{et al.} \cite{ovsjanikov2008gis} proposed detecting {\em intrinsic symmetries} of a shape by
 analyzing the eigenfunctions of the Laplace-Beltrami operator.
Intrinsic symmetry is manifested in the existence of a self-isometry $f : X \rightarrow X$, under which the
 metric structure of the shape is preserved, i.e., $d = d \circ (f \times f)$, where $d$ is the commute time metric. % \cite{raviv07sym}.
Ovsjanikov \emph{et al.} \cite{ovsjanikov2008gis} observe that for any intrinsic reflection symmetry $f$,
simple eigenfunctions of $\Delta$ satisfy $\phi_i \circ f = \pm \phi_i$. Thus, the symmetries of $X$ can be
  parameterized by the \emph{sign signature} $(s_1,s_2,\hdots)$; $s_i \in \{-1,1\}$ such that
  $\phi_i \circ f = s_i \phi_i$.

%The symmetries of $X$ are detected by testing different sign signatures of the first few eigenfunctions.
%
Given a truncated sign signature $(s_1,\hdots,s_K)$, define an energy
\begin{eqnarray*}
E(s_1,\hdots,s_K) = \int \min_{x' \in X} \sum_{i=1}^K \frac{1}{\lambda_i} (s_i \phi_i(x) - \phi_i(x'))^2 da.
 %\| \Phi_s(x) - \Phi(x') \|_2^2.
\end{eqnarray*}
It is easy to show that $E = 0$ for sign signatures corresponding to intrinsic symmetries, and $E\approx 0$ for
 approximate symmetries satisfying $d \approx d \circ (f \times f)$.
% and will be. %.\footnote{This straightforward symmetry detection approach has complexity exponential in $d$.
% For large $d$, Ovsjanikov \emph{et al.} \cite{ovsjanikov2008global} propose a fast heuristic.}
%
The symmetry itself is recovered as
\begin{eqnarray*}
f(x) &=& \mathop{\mathrm{argmin}}_{x' \in X} \sum_{i=1}^K \frac{1}{\lambda_i} (s_i \phi_i(x) - \phi_i(x'))^2.
 %\| \Phi_s(x) - \Phi(x') \|_2.
\end{eqnarray*}
Employing our equi-affine-invariant Laplacian, the detection of intrinsic symmetries can be made under
 affine transformations of the shape.

%{\bf Ovsjanikov's symmetries are in the sense of the COMMUTE TIME metric. This means that in our version, the result will be truly
%affine-invariant. We need to extend this point and demonstrate it in a figure!}

\begin{figure}[t!]
%\begin{minipage}[b]{1\linewidth}
  \centering \includegraphics*[width=.7\columnwidth]%,bb=1 1 1230 720]
            {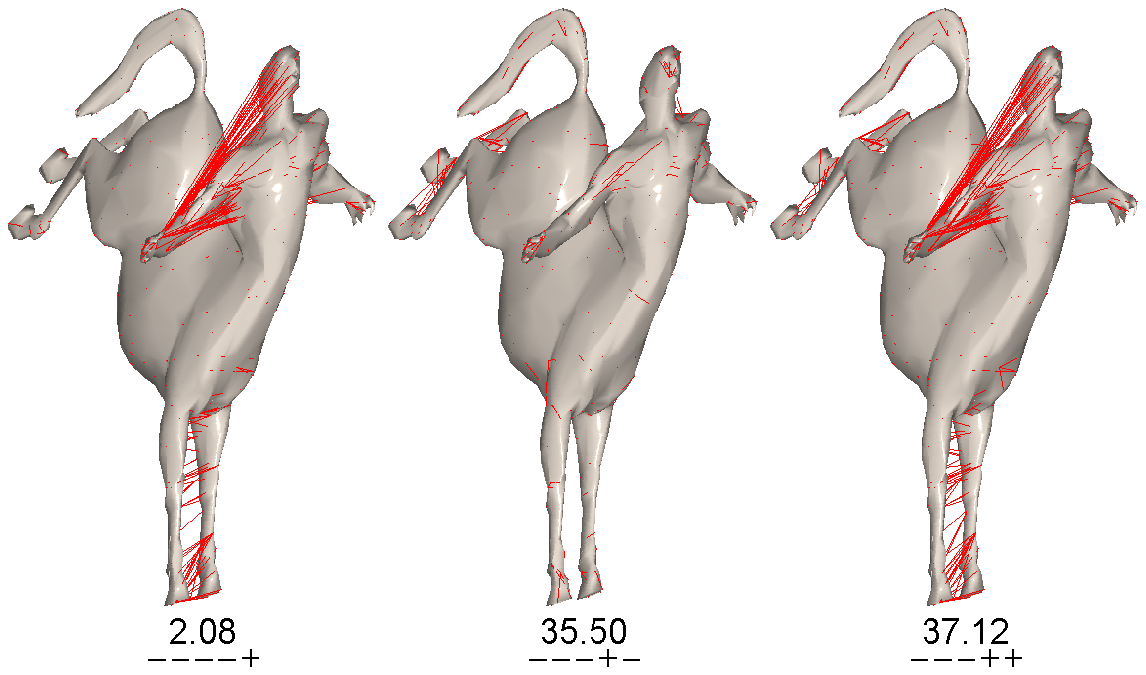}\\
%\end{minipage}\\
%
%\begin{minipage}[b]{1\linewidth}
  \centering \includegraphics*[width=.7\columnwidth]% ,bb=1 1 1230 720]
            {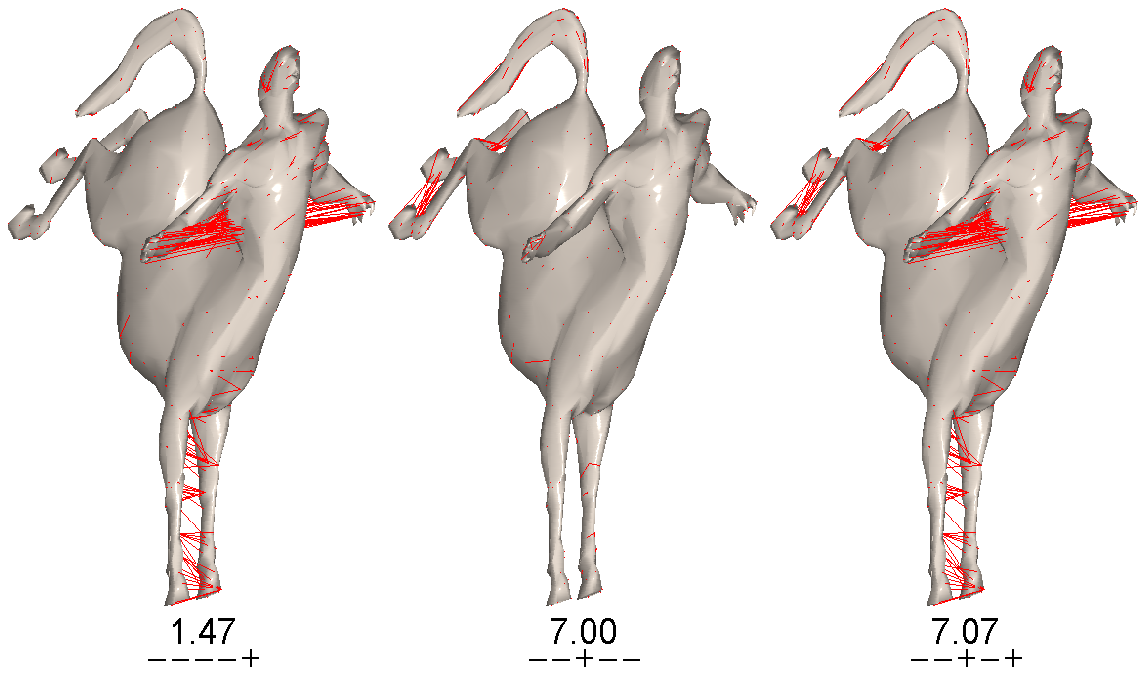}
%\end{minipage}\\
   \caption{\label{fig:sym} \small{Symmetry detection with the method of \cite{ovsjanikov2008gis} using standard (first row) and the proposed affine-invariant (second row) Laplace-Beltrami operator. Red lines depict the point correspondence $f$. For each sign signature, the corresponding error $E$ is shown.} }
%    \end{center}
\end{figure}

Figure~\ref{fig:sym} shows an example of intrinsic symmetry detection with the method of \cite{ovsjanikov2008gis} using standard (first row) and the proposed affine-invariant (second row) Laplace-Beltrami operator
on an intrinsically symmetric centaur shape that underwent a mild isometric and affine transformation.
In this experiment, we use the five first non-trivial eigenfunctions and show the approximate symmetries corresponding to the sign signatures producing the smallest values of $E$
(sign signature $+++++$ corresponding to the identity transformation was ignored).
While no meaningful symmetries are detected in the first case, using our affine-invariant Laplace-Beltrami operator we were able to detect three approximate symmetries of the shape: hands and forward legs reflection (left), rear legs reflection (center), and full body reflection (right).

\section{Conclusions}
\label{sec:concl}

We introduced an equi-affine-invariant Laplace-Beltrami
operator on two-dimensional surfaces, and showed that it can be
utilized to construct affine-invariant local and global diffusion
geometric structures.
Performance of the proposed tools was demonstrated on shape retrieval,
correspondence, and symmetry detection applications.
Our results show that affine-invariant diffusion geometries gracefully compete
with, and sometimes even outperform, their classical counterparts
under isometric changes and in the presence of geometric and topological noise,
while significantly outperforming the latter under affine transformations.
%

% \paragraph{Limitations.} A limitation of our approach is the fact that it gives only equi-affine rather that general affine invariance. We, however,
% demonstrated that either by considering the scale-invariant version of HKS or the commute time distance still allows the construction of
% both local and global affine-invariant structures. Another limitation is the fact that our method works only in cases where the embedding
% co-dimension of the surface is greater than zero (e.g., a two dimensional surface embedded in $\RR^3$, as in the studied examples). 
% In future studies, we plan to extend our construction to planar and volumetric shapes having embedding co-dimension $0$.

Extension of the proposed equi-affine framework into fully affine invariance (including scale)
could be accomplished by either exploiting the scale invariance property of the commute time
distance, or the consideration of scale invariant signatures, two approaches we plan to
study in  the future.

%We used the affine invariant metric for surfaces
% to extended the field of diffusion geometry.
%The eigenfunctions of the resulting affine invariant Laplacian
% operator serve as building blocks in constructing  invariant
% signatures and diffusion distances.
%With theses tools we have been able to compete with state of
% the art methods in shape matching and symmetry analysis and
% extend, if only just a little, the set of deformations we
% can deal with while trying to understand the structure of
% shapes.
%
%

\section{Acknowledgements}
This research was supported in part by The Israel Science Foundation (ISF) grant number 623/08, 
and by the USA Office of Naval Research (ONR) grant.

{\small
\bibliographystyle{ieee}
\bibliography{bib_art}

\begin{thebibliography}{10}\itemsep=-1pt

\bibitem{Axioms:Image:Proc:morel93}
L.~Alveraz, F.~Guichard, P.~L. Lions, and J.~M. Morel.
\newblock Axioms and fundamental equations of image processing.
\newblock {\em Archive for Rational Mechanics and Analysis}, 123(3):199--257,
  1993.

\bibitem{AryaMount:98:ICP}
S.~Arya, D.~M. Mount, N.~S. Netanyahu, R.~Silverman, and A.~Y. Wu.
\newblock An optimal algorithm for approximate nearest neighbor searching.
\newblock {\em J. ACM}, 45:891--923, 1998.

\bibitem{berard}
P.~B{\'e}rard, G.~Besson, and S.~Gallot.
\newblock Embedding riemannian manifolds by their heat kernel.
\newblock {\em Geometric and Functional Analysis}, 4(4):373--398, 1994.

\bibitem{Proj:Inv:Bruck97}
A.~M. Bruckstein and D.~Shaked.
\newblock On projective invariant smoothing and evolutions of planarcurves and
  polygons.
\newblock {\em Journal of Mathematical Imaging and Vision}, 7:225--240, June
  1997.

\bibitem{Buchin1983AffineGeometry}
S.~Buchin.
\newblock {\em Affine differential geometry}.
\newblock Beijing, China: Science Press, 1983.

\bibitem{Best:Scale:Space:Caselles:96}
V.~Caselles and C.~Sbert.
\newblock What is the best causal scale space for 3d images?
\newblock {\em SIAM Journal Applied Math.}, (56):1196--1246, 1996.

\bibitem{diff}
R.~R. Coifman and S.~Lafon.
\newblock Diffusion maps.
\newblock {\em Applied and Computational Harmonic Analysis}, 21:5--30, July
  2006.

\bibitem{Dziuk88}
G.~Dziuk.
\newblock {Finite elements for the Beltrami operator on arbitrary surfaces}.
\newblock In S.~Hildebrandt and R.~Leis, editors, {\em Partial differential
  equations and calculus of variations}, pages 142--155. 1988.

\bibitem{dinosaurs}
D.~Ghosh, N.~Amenta, and M.~Kazhdan.
\newblock Closed-form blending of local symmetries.
\newblock In {\em Proc. SGP}, 2010.

\bibitem{gro:GEOMETRY}
M.~Gromov.
\newblock {\em Structures {M}{\'e}triques {P}our les {V}ari{\'e}t{\'e}s
  {R}iemanniennes}.
\newblock Number~1 in Textes Math{\'e}matiques. 1981.

\bibitem{levy2006lbe}
B.~L{\'e}vy.
\newblock {Laplace-Beltrami} eigenfunctions towards an algorithm that
  ``understands'' geometry.
\newblock In {\em Proc. Shape Modeling and Applications}, 2006.

\bibitem{MS_GMOD}
M.~Mahmoudi and G.~Sapiro.
\newblock Three-dimensional point cloud recognition via distributions of
  geometric distances.
\newblock {\em Graphical Models}, 71(1):22--31, January 2009.

\bibitem{MHKCB}
D.~Mateus, R.~P. Horaud, D.~Knossow, F.~Cuzzolin, and E.~Boyer.
\newblock Articulated shape matching using laplacian eigenfunctions and
  unsupervised point registration.
\newblock {\em Proc. CVPR}, June 2008.

\bibitem{mem:sap1:GEOMETRY}
F.~M{\'e}moli and G.~Sapiro.
\newblock A theoretical and computational framework for isometry invariant
  recognition of point cloud data.
\newblock {\em Foundations of Computational Mathematics}, 5:313--346, 2005.

\bibitem{cordelia}
K.~Mikolajczyk, T.~Tuytelaars, C.~Schmid, A.~Zisserman, J.~Matas,
  F.~Schaffalitzky, T.~Kadir, and L.~V. Gool.
\newblock A comparison of affine region detectors.
\newblock {\em IJCV}, 65(1--2):43--72, 2005.

\bibitem{Geometric:Evolutions:Olver:97}
P.~Olver, G.~Sapiro, and A.~Tennenbaum.
\newblock Invariant geometric evolutions of surfaces and volumetric smoothing.
\newblock {\em SIAM Journal Applied Math.}, (57):176--194, 1997.

\bibitem{BroBroOvsGui09}
M.~Ovsjanikov, A.~M. Bronstein, M.~M. Bronstein, and L.~J. Guibas.
\newblock Shape {G}oogle: a computer vision approach to invariant shape
  retrieval.
\newblock In {\em Proc. NORDIA}, 2009.

\bibitem{ovsjanikov2008gis}
M.~Ovsjanikov, J.~Sun, and L.~J. Guibas.
\newblock {Global intrinsic symmetries of shapes}.
\newblock In {\em Computer Graphics Forum}, volume~27, pages 1341--1348, 2008.

\bibitem{reuter2009discrete}
M.~Reuter, S.~Biasotti, D.~Giorgi, G.~Patan{\`e}, and M.~Spagnuolo.
\newblock {Discrete Laplace--Beltrami operators for shape analysis and
  segmentation}.
\newblock {\em Computers \& Graphics}, 33(3):381--390, 2009.

\bibitem{reuter}
M.~Reuter, F.-E. Wolter, and N.~Peinecke.
\newblock {L}aplace-spectra as fingerprints for shape matching.
\newblock In {\em Proc. ACM Symp. Solid and Physical Modeling}, pages 101--106,
  2005.

\bibitem{rustamov2007lbe}
R.~M. Rustamov.
\newblock {Laplace-Beltrami eigenfunctions for deformation invariant shape
  representation}.
\newblock In {\em Proc. SGP}, pages 225--233, 2007.

\bibitem{Sochen04}
N.~Sochen.
\newblock Affine-invariant flows in the {B}eltrami framework.
\newblock {\em Journal of Mathematical Imaging and Vision}, 20(1):133--146,
  2004.

\bibitem{sunHKS}
J.~Sun, M.~Ovsjanikov, and L.~J. Guibas.
\newblock A concise and provably informative multi-scale signature based on
  heat diffusion.
\newblock In {\em Proc. SGP}, 2009.

\end{thebibliography}
}

\end{document}